# Force-based Algorithm for Motion Planning of Large Agent Teams

Samaneh Hosseini Semnani, Anton de Ruiter, Hugh Liu.

*Abstract*—This paper presents a distributed, efficient, scalable and real-time motion planning algorithm for a large group of agents moving in 2 or 3-dimensional spaces. This algorithm enables autonomous agents to generate individual trajectories independently with only the relative position information of neighboring agents. Each agent applies a force-based control that contains two main terms: collision avoidance and navigational feedback. The first term keeps two agents separate with a certain distance, while the second term attracts each agent toward its goal location. Compared with existing collision-avoidance algorithms, the proposed force-based motion planning (FMP) algorithm is able to find collision-free motions with lower transition time, free from velocity state information of neighbouring agents. It leads to less computational overhead. The performance of proposed FMP is examined over several dense and complex 2D and 3D benchmark simulation scenarios, with results outperforming existing methods.

*Index Terms*— collision avoidance, distributed algorithms, evolutionary computation, flocking algorithm, motion planning, multi-agent systems, trajectory optimization

## I. Introduction

Motion planning is a significant problem in multi-agent systems. This problem first concerns finding a one-to-one assignment between a set of given agents' initial locations and the corresponding destination locations and second finding the continuous trajectories connecting each agent's initial location to its goal location. Using motion functions, the agents should be able to reach their goal destinations without colliding with each other or environment obstacles. In addition, the motion planning algorithm should be able to satisfy constraints such as the maximum velocity of the agents and the minimum required separation distance between them. These constraints are dependent on the agents' dynamic limitations and may vary from one application to another.

This problem has attracted much interest in the research community and various motion planners are developed, from discrete planning, motion coordination algorithms, to real-time collision avoidance methods. However, to the authors' best knowledge, no existing solution satisfies all requirements of applications simultaneously such as robustness, scalability, optimality, low computational and communication overhead, being fast, real-time and flexible, etc. In this paper, we present an algorithm that attempts to balance these goals.

The main contribution of this paper is adapting the flocking algorithm, which has recently shown outstanding behavior in the field of mobility control [1-2], to solve the motion planning problem for a large group of agents while satisfying the above-mentioned multiple requirements of real applications. This adaptation, presented as the Force-based motion planning (FMP) algorithm, requires essential modification of the flocking algorithm, including elimination of velocity matching and flock centering terms, modification of the collision avoidance term to generate collision-free motions with required separation distance as well as addition of multi-target navigational feedback to guide agents toward their goal locations. Optimal reciprocal collision avoidance (ORCA) is one well-received real-time collision avoidance algorithm that has recently gained significant attention as a fast and scalable tool. ORCA, while claiming to find an optimal algorithm, struggles with finding a time-optimal solution. Moreover, it requires position and velocity to be exchanged over a communication network among the agents. The presented FMP algorithm is able to find collision free paths with lower transition time without requiring the velocity information of neighbors and with less computational overhead in comparison with ORCA. The performance of FPP is examined over several dense and complex 2D and 3D benchmark simulation scenarios. The results show that FPP computes collision-free and deadlock-free paths for all the scenarios while outperforming ORCA with respect to both transition and execution time. Our approach can compute safe and smooth trajectories for thousands of agents in dense environments with static and dynamic obstacles in few milliseconds.

The outline of the rest of the paper is given as follows. Section II discusses the main research in the field of agents' motion planning and their drawbacks, in addition to the Flocking algorithm and its extensions as the base of the solution proposed in this paper. Section III presents FMP, the proposed force-based algorithm for agents' motion planning and the mathematical analyses on its computation complexity, its ability to provide collision free motions and guarantee a minimum separation distance and the convergence of the algorithm. Section IV demonstrates the simulation results. Finally, concluding remarks and recommendations for future directions are given in Section V.

## II. Related Work

*A. Motion Planning algorithms*

Multi-agent motion planning problem has attracted much interest in the research community and various motion planners have been developed. Existing work are traditionally classified into two categories, centralized and distributed approaches, however there are some approaches that cannot fit in these two categories.

Centralized approaches [3-6] are suitable for small problems and are able to benefit from many standard optimization algorithms, however, they are often computationally expensive especially in large problems and are not suitable for large teams. A simple centralized approach is motion coordination. Motion coordination refers to controlling the velocity of agents along their fixed pre-planned motions such that none of the them pass intersecting points between their motions at the same time. This approach introduces a configuration space and represents agent-agent collisions as obstacles in this space [3]. Efforts have been done to reduce the computational complexity of the calculations [5] but still for large-scale problems the computational complexity is much more than the distributed algorithms.

"This work was supported by the Ontario Centers of Excellence, grant number 27481."

S. Hosseini Semnani is with the Aerospace Engineering Department, Ryerson University, Toronto, Canada (e-mail: samaneh.hoseini@ryerson.ca).

A. de Ruiter is with the Aerospace Engineering Department, Ryerson University, Toronto, Canada (e-mail: aderuiter@ryerson.ca).

H. Liu is with the Faculty of Applied Science & Engineering, University of Toronto, Toronto, Canada (e-mail: liu@utias.utoronto.ca)

Distributed approaches [7-8] use local, reactive collision avoidance algorithms to let each agent plan its motion independently and resolve conflicts in real-time when impending collisions are detected. Distributed approaches are highly scalable as each agent makes decisions independently and their robustness against disturbance is appealing as are doing real-time computations. However, they do not provide any means to optimize the trajectories from time or energy perspective. Optimal reciprocal collision avoidance (ORCA) [7] is one well known algorithm in this group. Given perfect knowledge of neighboring positions and speeds, the ORCA algorithm lets each agent compute its optimal velocity independently by solving an optimization problem at each time step. ORCA requires relative position and velocity to be shared between neighboring agents [9].

Another approach is to solve trajectory planning problems in multiple stages [10-12]. The algorithms that use this method often require solving one large optimization problem in one of their stages in which the decision variables define all of the agents' trajectories. As a result, they do not scale well to large teams. In [10], the authors decouple the problem into a discrete planning stage followed by a refinement stage. In this method, a discrete planning algorithm is applied for both finding an assignment between agents' initial and final locations and for computing a sequence of waypoints. This discrete planner solves the problem by reduction to a maximum-flow problem. Then continuous refinement methods start working on the produced discrete plan to compute smooth trajectories. However, such an algorithm is unable to avoid dynamic obstacles as it is not a real-time algorithm and plans the trajectories beforehand. It also requires heavy calculations especially in large size problems which results in low speed (e.g., on the order of minutes) for a team with 32 agents while using a CPU with 10 cores [10].

An alternative approach is to use Voronoi diagrams and the Lloyd algorithm for planning collision-free trajectories. Although Voronoi diagrams were initially used for finding optimal environment coverage and agent placement, some papers [13] apply this technique to trajectory planning by iteratively calculating the Voronoi cells and planning trajectories to the point in the cell that is closest to the agent's goal position. The trajectories generated are naturally collision-free since the points only move within their cell at each iteration. However, in real distributed simulations especially in 3D environments, this method still requires calculation of Voronoi diagrams and the closest points at each iteration which imposes high computational complexity over the agents, especially when the number of agents and iterations to converge increase. In addition, this method does not optimize the agents' motions in terms of time or energy.

Another recent approach is the application of deep reinforcement learning for multi agent motion planning [14-15]. This approach offloads the expensive online computation (for predicting interacting patterns) by learning a policy and a value function applying reinforcement learning techniques that implicitly encodes cooperative behaviors. The learned policy will be used by each agent later in real-time to select best action at each time. The reinforcement learning is an active research topic at its early stage with many unanswered questions. Current development still has drawbacks e.g., the policy cannot be well generalized to new scenarios that do not appear in the entire training period. Also, it is not suitable for large-scale problems due to large computational complexity of the training part.

*B. Flocking Based Algorithms*

This section briefly introduces the Flocking algorithm [1] and its extensions that are related to agents' motion planning as the basis of the FMP algorithm.

Flocking is a distributed group behavior of a large number of autonomous interacting agents able to self-organize without having a leader or any other central control. This algorithm is based on three Reynolds rules [1]: flock centering, collision avoidance and velocity matching. Reference [2] provides a solid mathematical background for the three main Reynolds rules in which each agent applies a force vector $u_i$ as a control input unit. Equations (1) to (5) present vehicle dynamics and the $u_i$ force vector for agent $i$:

$$\dot{p}_i = v_i, \quad \dot{v}_i = u_i, \quad u_i = u_i^\alpha + u_i^\gamma \quad (1)$$

$$u_i^\alpha = f_i^g + f_i^d \quad (2)$$

$$f_i^g = \sum_{j \in N_i} \phi_\alpha(\|p_j - p_i\|_\alpha) n_{ij} \quad Gradient-based\ term \quad (3)$$

$$f_i^d = \sum_{j \in N_i} a_{ij}(q)(v_j - v_i) \quad Velocity-Consensus\ term \quad (4)$$

$$u_i^\gamma = -c_1(p_i - p_r) - c_2(v_i - v_r)\ c_1, c_2 > 0 \quad Navigational\ Feedback \quad (5)$$

Where $p_i, v_i \in \mathbb{R}^3$ denote the position and velocity of agent $i$ respectively. $p_r, v_r \in \mathbb{R}^3$ are position and velocity of a dynamic/static agent that represents a group objective and can be viewed as a moving target. $N_i$ is the set of neighbors of agent $i$. $a_{ij}(q)$ is an element of a spatial adjacency matrix $A(q)$. $n_{ij} = (p_j - p_i/\sqrt{1 + \epsilon \|p_j - p_i\|^2}$ is a vector along the line connecting $p_i$ to $p_j$ and $\epsilon \in (0,1)$ is a fixed parameter of the $\sigma$-norm. $c_1, c_2$ are two positive constant values. applies $\phi_\alpha$ is an Attractive/repulsive pairwise function depicted in Fig. 1, in which $d_\alpha$ is the preferred distance between the agents and $r_\alpha$ is the communication distance between them. $f_i^g$ attempts to keep all the agents' relative distances close to $d_\alpha$.

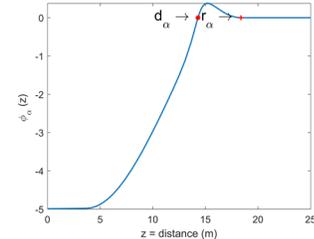

Fig. 1. Attractive/repulsive pairwise function used in gradient-based term of the Flocking algorithm.

The term $f_i^d$ is a damping force that tries to keep the velocity of each agent close to its flockmates velocities and $u_i^\gamma$ is the navigational feedback and attempts to attract all the agents toward a group objective. See [2] for more details of (3) to (5). Applying the flocking algorithm over a group of agents results in the creation of a network of the agents moving around a target $(p_r, v_r)$ while trying to keep distance between themselves equal to $d_\alpha$. Fig. 2 presents the result of flocking algorithm after a few seconds of its start time [2].

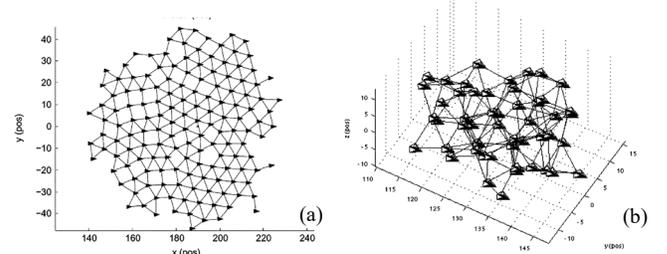

Fig. 2. Network of agents created by Flocking algorithm a few seconds after starting in (a): 2D and (b): 3D environments [2]

Semi-Flocking is a recent extension of the main flocking algorithm, which is designed for tracking multiple targets, each being a common objective for a sub-group of agents [16, 17]. Section III shows how we use this characteristic of the Semi-Flocking algorithm for motion planning of large agent teams. The Semi-Flocking algorithm changes the navigational-feedback term ($u_i^\gamma$) in the force function $u_i$ as shown in (6) while keeping the gradient-based and velocity-consensus terms unchanged.

$$u_i^\gamma = f_i^\gamma(p_i, p_{t1}, \ldots, p_{t1}, v_i, v_{t1}, \ldots, v_{tm}) =$$
$$\sum_{j=1}^{m} \varphi(p_{tj} - p_i) \cdot \left(c_{1j}(p_{tj} - p_i) + c_{2j}(v_{tj} - v_i)\right)/n_{tj} \quad (6)$$

where $m$ represents the number of targets, $c_{1j}$, $c_{2j}$ are positive constant values, $p_i$ and $v_i$ are position and velocity of sensor $i$ respectively. $p_{tj}$ and $v_{tj}$ are position and velocity of target $j$ respectively. $n_{tj}$ represents the number of sensors currently tracking target $tj$ and $\varphi$ is a switching function taking 0-1 values. See [16] for more details of (6). The result of applying the Semi-Flocking algorithm on a multi-target problem is small quasi α-lattices of agents around each target while still leaving some agents free in the area for searching new targets (see Fig. 9 in [16]).

### III. Approach

This section presents a Force-based motion planning (FMP) algorithm, a fast, real-time and effective solution for 2D and 3D motion planning based on the Flocking algorithm described in Section II.B. Target tracking is just one of the applications of the Flocking algorithm that is introduced in [2]. However, if the Flocking algorithm is able to guide the agents toward a target without any collision, we can exploit that in agents' motion planning. The initial idea is to put the target in a desired goal location and apply an adopted version of the Flocking algorithm to find collision-free motions from agents' initial locations to the goal location. Since we do not want to guide all the agents to one goal position, we need to increase the number of targets to be exactly equal to the number of agents ($m = n$) to have a separate pair of (initial, goal) locations for all the agents. Semi-Flocking presents a solution for increasing the number of targets from 1 in the Flocking algorithm to $m$ ($m < n$) [16]. The FMP algorithm continues this track to have $n$ targets where each is assigned to one agent. In addition, FMP makes essential changes in the control input force vector of flocking framework ($u_i$) including elimination of velocity consensus term and modification of the gradient based and navigational feedback terms.

*A. Problem statement*

Given n agents, each with its own target location, the objective is to guide them to their target locations, while maintaining a specified minimum distance between them (as a result of considering the agents volumes and localization errors), avoiding collisions with environmental obstacles, and satisfying communication constraints and maximum agent velocity constraints in horizontal and vertical directions.

*B. Assumptions*

The following assumptions have been made in this research:
- We have $n$ agents deployed in a $3D$ environment.
- Each agent is moving independently but in coordination with the other agents' motions. Let $p_i, v_i, u_i \in \mathbb{R}^3$ denote the position, velocity and acceleration of the agent $i$ respectively, $p = col(p_1, \ldots, p_n), v = col(v_1, \ldots, v_n)$. Agent $i$ obeys the following dynamics:
$$\begin{cases} \dot{p}_i = v_i \\ \dot{v}_i = u_i \end{cases} \quad 1 \leq i \leq n$$
- $r$ is radius of a neighborhood within which the agents must interact
- Each agent is able to measure the relative position of its neighboring agents and the closest point of obstacles.
- The set of initial and goal locations of the agents are given:
  - $\mathcal{I} = \{\mathcal{I}_1, \mathcal{I}_2, \ldots, \mathcal{I}_n\}$ : set of initial positions of agents
  - $\mathcal{F} = \{\mathcal{F}_1, \mathcal{F}_2, \ldots, \mathcal{F}_n\}$ : set of goal positions of agents
  where $\mathcal{I}_i, \mathcal{F}_i$ represent the initial and final position of agent i respectively.
- We have $n$ hypothetical static targets placed at the agents' goal positions. $\mathcal{T} = \mathcal{F}$
- Separation distance between agents in $\mathcal{I}$ and $\mathcal{F}$ is greater than or equal to $d$, where $d$ is an upper bound for required minimum separation distance between agents
- $d^*$: required minimum separation distance between agents ($d^* \leq d$).
- $v_{max}$: maximum allowable velocity of the agents in each direction is given

*C. Force-based Motion Planning (FMP) algorithm*

In contrast to the Semi-flocking algorithm which was designed for multi-agent target tracking, the FMP algorithm is a motion planning algorithm, therefore we need to modify the three main terms of the Flocking algorithm: gradient based, velocity consensus and navigational feedback to fit the requirement of this problem and to satisfy separation and maximum agent velocity constraints.

To reach this point, FMP first generates $n$ hypothetical targets and places them at the given goal locations. Each agent is uniquely assigned to one target. In this paper, it is assumed that all targets have been assigned. Hungarian assignment [18] is one possible method in which the global cost function is minimized (see (7)).

$$\min \sum_{i=1}^{n} \sum_{j=1}^{n} \|\mathcal{T}_j - \mathcal{I}_i\| \quad (7)$$

where $\|x\|$ is the Euclidean norm of $x$. A standard solution for this minimization problem is the so-called Hungarian algorithm [18] which presents $O(N^3)$ computational cost. Reference [19] introduces a distributed version of this algorithm which makes it suitable for distributed problems as studied in this paper. The Auction algorithm [20] and its distributed version [21] are lower cost alternatives to the Hungarian algorithm which produce suboptimal solutions in significantly less time.

After target-agent assignment, each agent $i$ applies the control function presented in (8) iteratively until it reaches to its goal location.

$$u_i = \begin{cases} 0 & \text{if } v_i^T \bar{u}_i > 0 \text{ and } \|v_i\| \geq V_{max} \\ \bar{u}_i = f_i^{\mathcal{R}} + f_i^{\gamma} & \text{otherwise} \end{cases} \quad (8)$$

where $f_i^{\mathcal{R}}$ is the repulsive force and $f_i^{\gamma}$ is the navigational feedback that attracts the agents towards the assigned targets. Equations (9) and (10) describe how $f_i^{\mathcal{R}}$ and $f_i^{\gamma}$ are calculated.

$$f_i^{\mathcal{R}} = \sum_{j \in N_i} \left( \phi(\|p_j - p_i\|) \frac{p_j - p_i}{\|p_j - p_i\|} \right) \quad \text{Repulsive Function} \quad (9)$$

$$f_i^{\gamma} = -c_1(p_i - \mathcal{T}_i) - c_2(v_i) \quad \text{Navigational Feedback} \quad (10)$$

$$\phi(z) = \begin{cases} -\rho \times (z - r)^2 & 0 < z < r \\ 0 & z \geq r \end{cases} \quad (11)$$

$$N_i = \{j \in \mathcal{V} - \{i\}: \|p_j - p_i\| < r\} \quad \mathcal{V} = \text{the set of indices of agents} \quad (12)$$

where $\phi$ is a repulsive function defined in (11) and depicted in Fig. 3(a) and $N_i$ represents the agents within a neighborhood of agent $i$ and is defined in (12). $c_1, c_2$ are two positive constant values, $\rho$ is a positive repulsive gradient and $r$ is the communication radius. Section C.1 explains how we handled agents' maximum velocity constraint in the $u_i$ function.

The control function in FMP, $u_i$, has three main differences with the Flocking control function (1). First, the velocity-consensus term is omitted here. The main functionality of velocity-consensus term is matching the velocity of each agent with nearby flockmates. However, when the Flocking algorithm is going to be used for motion planning of individual agents each one following a separate target, there is no need to match the velocity of each agent with its neighbors.

Second, the attractive/repulsive function of the flocking algorithm (3) is changed to an only repulsive function presented in (9). As described in Section II.B attractive/repulsive function in the gradient term of the Flocking algorithm creates a spring model that attempts to keep the distance between two agents equal to a predefined value $d$. This function is useful in the Flocking algorithm because it needs to create a network of agents placed at a preferred distance from each other. However, in the FMP algorithm we only need to keep the agents strictly apart from each other and there is no need to keep them at a specific distance of one another. Therefore, we only need to have a repulsive force between each two agents to repel them from one another when they become close. This function is defined in (11) and depicted in Fig. 3(a). The value of the parameter $\rho$ in this function should be selected such that the function be able to provide a high enough repulsive force. In this paper, we take $\rho = 7.5 \cdot 10^6$.

As represented in Fig. 3(a), two agents start repelling each other when they come within a given communication distance of one another ($r$). The intensity of repelling force depends on the distance between two agents. Section C.1 describe how the communication distance $r$ is calculated based on the given separation distance $d$ and maximum allowed velocity $V_{max}$. Equation (13) shows the smooth potential function between two agents, based on their repulsive force (11). This function is depicted in Fig. 3(b).

$$\psi(z) = \int_0^z \phi(s)ds = \begin{cases} -(\rho/3) \times (z-r)^3 + C & 0 < z < r \\ C & z \geq r \end{cases} \quad (13)$$

where C is a constant value. This repulsive potential function induces a collective potential function in the form:

$$V(p) = (1/2) \sum_i \sum_{j \neq i} \psi(\|p_j - p_i\|) \quad (14)$$

Comparing Fig. 1 and Fig. 3(a) reveals the difference between the attracting/repulsive force in the main Flocking algorithm and the only repulsive force in the FMP algorithm. It is clear that the attractive force in the Flocking algorithm in $[d, r]$ is replaced with a repulsive force in the FMP algorithm. It also provides a much higher repulsive force in $[0, d]$ interval which significantly pushes two agents apart when they want to become closer than $d$ and guarantee our required minimum separation distance $d^*$. Theoretical analysis in Section III.D.2 and simulation results presented in Section IV demonstrate this fact.

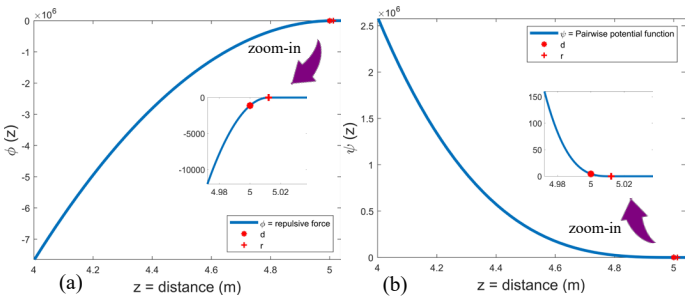

Fig. 3. Repulsive function (a) and Smooth pairwise potential function $\psi(z)$ (b) used in FMP algorithm for $d = 5$ in $[4, 5.03]$ interval

The third and final difference between the control functions of the Flocking and FMP algorithms is about the navigational feedback term. The navigational term of FMP is very similar to the navigational term of the Semi-Flocking that is described in Section II.B, as they both attract the agents toward multiple targets. However, there are still essential differences between them in terms of the number of agents assigned to track each target. Flocking and Semi-Flocking algorithms are both target tracking algorithms that require to have a group of agents around target/targets to provide suitable target coverage. While the main objective of the FMP algorithm is to create collision free motions toward the targets.

*C.1 Maximum velocity constraint*

The agents' velocity restrictions are the other constraints in a real motion planning problem which should be considered in the proposed algorithm. A motion planning algorithm may impose a velocity to the agents which is beyond their physical ability. This constraint is explicitly considered in the FMP algorithm. The velocity of each agent is influenced by its control input $u_i$. Therefore, to control the velocity, the FMP algorithm should restrict the two force functions of $\bar{u}_i$ in Equation (8) such that they don't cause over limit velocities. We restrict the value of the repulsive force by selecting a communication distance $r$ that is very close to $d$. Equation (15) shows how we calculate $r$:

$$r = \sqrt[3]{3v_{max}^2/2\rho} + d \quad (15)$$

To keep the velocity always lower than $V_{max}$ after calculating $\bar{u}_i$, a capping function (cap_velocity(i) in Algorithm 1) is added to the algorithm that limits each agent's velocity to its maximum value. It is equivalent to the restriction condition used in the first term of Equation (8). Maximum velocity can be defined in each direction separately ($v_{max-x}, v_{max-y}, v_{max-z}$). However, for simplicity we assume maximum velocity in all the directions are equal to $V_{max}$.

Algorithm 1 is a sampled data implementation of the continuous-time FMP algorithm discussed thus far. As illustrated in this pseudocode, each agent calculates its next velocity and position at each time step in a completely distributed manner without requiring any central control unit. It only requires the knowledge of relevant location of its neighbors.

---

**Algorithm 1: Force-based Motion Planning (FMP) algorithm**

**Input:**
- $n$ = number of agents (= number of targets)
- $\mathcal{I}$ = Initial position of agents  $\mathcal{F}$ = final position of agents
- $d^*$ = required minimum separation distance between agents
- $v_{max}$ = maximum velocity of the agents

**Output:**
- $\forall_{i=0}^n p_i(t)$    $p_i(t): (p_{ix}(t), p_{iy}(t), p_{iz}(t))$ = position of agent $i$ at time $t$
- $\forall_{i=0}^n v_i(t)$    $v_i(t): (v_{ix}(t), v_{iy}(t), v_{iz}(t))$ = velocity of agent $i$ at time $t$
  ($0 \leq t$ increases in $\Delta t$ time intervals)

**begin**
  $v_i(0) = (0,0,0)$              // initialize agent velocity
  $\mathcal{A}$ = agent-target assignment     // e.g. Hungarian
  $\mathcal{T}_i = \mathcal{F}_{\mathcal{A}_i}$              // initialize target position
  Calculate $d$ and $r$ using (37) and (15)
  **for** ($t = 0$;  $t < MaxSimTime$;  $t +\!\!= \Delta t$)
    **for** $0 <= i < n$ **do in parallel**
      move_this_agent_to_new_position (i)
    **end for**
    MaxDis= agents_maximum_distanse_from_$\mathcal{F}$ ()
    **if** ( $MaxDis < EndMaxDis$ ) **do**
      break
    **end if**
  **end for**
**end**

---

**move_this_agent_to_new_position (i) {**
  $f_i^{\mathcal{R}}$ = repulsive_force (i),   $f_i^{\gamma}$ = navigational_feedback (i)
  $\bar{u}_i = f_i^{\mathcal{R}} + f_i^{\gamma}$
  $v_i(t) = v_i(t - \Delta t) + \bar{u}_i \Delta t$
  cap_velocity (i)                  //A hard constraint on velocity

```
    p_i(t) = p_i(t − Δt) + v_i(t)Δt}
```
---
```
repulsive_force (i) {
    f = 0
    for  0 <= j < n  do
        dist = ‖p_j − p_i‖
        if (dist < r and j ≠ i) then
            ForceComponent = −ρ × (dist − r )²
            f+= ForceComponent × (p_j − p_i)/‖p_j − p_i‖
        end if
    end for
return  f}
```
---
```
navigational_feedback (i) {
    f = −c₁(p_i − T_i) − c₂(v_i )
return  f}
```
---
```
cap_velocity(i) {
    if (‖v_i‖ > v_max) then
        v_i = (v_i/‖v_i‖) × v_max
    end}
```

*D. FMP with obstacle avoidance*

In this section, we present an extension of the FMP algorithm which is able to avoid multiple dynamic or static obstacles. The main idea is to use an agent-based representation of both dynamic and static obstacles by creating new species of agents called obstacle-agents. This approach is partially motivated by the work of Olfati-Saber [2]. Each obstacle-agent is a virtual agent that is induced by regular agents when they are in close proximity of an obstacle. The obstacles are spheres with a defined center and radius $(\hat{p}_j, \hat{r}_j)$. Fig. 4 depicts an obstacle $O_j$ which is in the neighborhood of agent $i$. As depicted in this figure, each obstacle has three parameters $\hat{p}_i, \hat{v}_i, \hat{r}_i$ which represent the position and velocity of the center of the obstacle and its radius. Also, $\acute{r}$ represents interaction range of an agent with neighboring obstacle-agents.

The interaction between regular agents and obstacle-agents is described using the notation $f_i^{obstacle}$ in (17). This equation is the same as (8) except it has additional repulsive force ($f_i^{obstacle}$) to avoid obstacles. The modified control force on agent $i$ is given by

$$u_i = \begin{cases} 0 & if\ v_i^T \bar{u}_i > 0\ and\ \|v_i\| \geq V_{max} \\ \bar{u}_i = f_i^{\mathcal{R}} + f_i^{\gamma} + f_i^{obstacle} & otherwise \end{cases} \quad (16)$$

$$f_i^{obstacle} = \sum_{j \in N_i^{obstacle}} \left(\phi_{obstacle}(\|\hat{p}_j − \hat{r}_j − p_i\|) \times \frac{\hat{p}_j − p_i}{\|\hat{p}_j − p_i\|}\right) \quad (17)$$

$$\phi_{obstacle}(z) = \begin{cases} -\acute{\rho} \times (z − \acute{r})^2 & 0 < z < \acute{r} \\ 0 & z \geq \acute{r} \end{cases} \quad (18)$$

$$N_i^{obstacle} = \{j \in \mathcal{V}_{obstacle}: \|\hat{p}_j − \hat{r}_j − p_i\| < \acute{r}\}$$

$$\mathcal{V}_{obstacle} = \text{the set of indices of obstacles} \quad (19)$$

where $f_i^{obstacle}$ is a repulsive function defined in (17) and $\acute{\rho}$ is its positive repulsive gradient and determines the severity of avoiding the obstacles. Here, we choose $r > \acute{r}, \rho = \acute{\rho}$, but in general, they can be chosen independently. $N_i^{obstacle}$ represents the set of neighboring obstacles of agent $i$ and is defined in (19). The other parameters are as defined in (8) to (12). Applying (16), agents are able to avoid obstacles in exactly the same way that they are able to avoid other agents. When the agents are getting close to an obstacle, they will receive a repulsive force $f_i^{obstacle}$ from the relevant virtual obstacle-agent. This repulsive force is the key element in collision avoidance ability of the FMP algorithm. In terms of sensing requirements, we assume that every agent is equipped with range sensors that allow the agent to measure the relative position between the closest point on an obstacle and itself $(\hat{p}_j − \hat{r}_j)$. Both radars and laser radars can be used as range sensors.

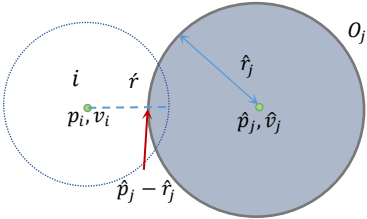

Fig. 4.  representation of a spherical obstacle in the neighborhood of an agent

*E. Theoretical Analysis*

This Section presents theoretical analysis on the computational complexity of the FMP algorithm. In addition, theoretical guarantees on collision free trajectories and the minimum separation distance between agents are given. Finally, mathematical analysis that proves the convergence of the algorithm are presented.

*D.1 Computational Complexity*

In this section, we calculate the computational complexity of each step of the FMP algorithm. Each agent at each step of the algorithm calculates two main forces: 1-repulsive force and 2-navigational force (See Algorithm 1). Calculation of the repulsive force only includes computing the value of the repulsive function for each neighbor. Assuming that each agent has found the $k$ neighboring agents in its vicinity, it takes $\mathcal{O}(k)$ time for FMP algorithm to calculate repulsive forces for each agent. Of course, in practice as the value of $r$ is selected very close to $d$, the number of each agents neighbors $(k)$ remains zero until it reaches to a distance from others $r$ which, as represented in Equation (15), is a little bit more than the value $d$. Calculation of navigational feedback force and cap_velocity() function are the two remaining computational steps for each agent at each iteration. The order of complexity of these two steps is $\mathcal{O}(1)$, therefore does not increase the total order of computational complexity of the algorithm. As a result, the total complexity of the algorithm for each agent at each iteration is $\mathcal{O}(k)$ and for all the agents at each iteration is $\mathcal{O}(n.k)$. However, in practice the execution time of the FMP is much lower than other algorithms with the same computational complexity, e.g. ORCA. Simulation results in Section IV confirm this observation. This is because calculation of the repulsive function for each neighbor is the only effective computational effort of each agent at each iteration in the FMP algorithm. Also, the number of neighbors $(k)$ is smaller in this algorithm. It is also the result of a lower number of steps each agent takes to reach to their goals using the FMP algorithm. It is worth mentioning that as the algorithm is completely distributed, and each agent is independent, the computations can be fully parallelized which decreases the execution time greatly.

*D.2 Collision free analysis*

This section presents analysis that shows using the FMP algorithm guarantees a minimum separation distance $d^*$ between the agents.

*D.2.1 Collision free proof*

Before presenting the minimum separation guarantee proof we need to define the structural dynamics corresponding to the FMP algorithm in addition to the Hamiltonian energy of the system. This analysis is performed with the continuous-time version of the algorithm and we know that the discretization approximates the continuous-time dynamics. The vehicle dynamics of each agent in the system is in the form:

$$\Sigma: \begin{cases} \dot{p}_i = v_i \\ \dot{v}_i = u_i \end{cases}, \quad u_i = \begin{cases} 0 & \text{if } v_i^T \bar{u}_i > 0 \text{ and } \|v_i\| \geq V_{max} \\ \bar{u}_i & \text{otherwise} \end{cases}$$
$$\bar{u}_i = -\partial U_\lambda(p)/\partial p_i - c_2 v_i \quad (20)$$

where $U_\lambda(p)$ is the aggregate potential function and is defined in (21). Solutions of (20) are understood to be in the Filippov sense.

$$U_\lambda(p) = V(p) + \lambda J(p) \quad (21)$$

where $V(p)$ is the collective potential function defined in (14), $\lambda = c_1 > 0$ and $J(p)$ is the moment of inertia of all the particles, defined as

$$J(p) = (1/2)\sum_{i=1}^{n}\|p_i - \mathcal{T}_i\|^2 \quad (22)$$

The structural Hamiltonian of system $\Sigma$ is as follows:

$$H_\lambda(p,v) = U_\lambda(p) + K(v) \quad (23)$$

where $K(v) = (1/2)\sum\|v_i\|^2 = (1/2)v^T v$ is the kinetic energy of the particle system.

**Claim 1**: If $\|v_i(0)\| \leq v_{max}$ then $\|v_i(t)\| \leq v_{max} \quad \forall t \geq 0$

**Proof**: Consider the kinetic energy of agent $i$: $K(v_i) = (1/2)v_i^T v_i$, then, $\dot{K}(v_i) = v_i^T \dot{v}_i = v_i^T u_i$. On the level set $\|v_i\| = v_{max}$ ($K(v_i) = (1/2)v_{max}^2$) we have:

$$\dot{K}(v_i) = \begin{cases} 0 & \text{if } v_i^T \bar{u}_i > 0 \\ v_i^T \bar{u}_i \leq 0 & \text{otherwise} \end{cases} \quad (24)$$

To show that $v_i(t)$ cannot leave the set $\|v_i(t)\| \leq v_{max}$, we only need to evaluate $\dot{K}(v_i)$ on the boundary of that set where $\|v_i(t)\| = v_{max}$ in which we have $\dot{K} = 0$ if $v_i^T \bar{u}_i > 0$ and $\dot{K} = v_i^T \bar{u}_i$ if $v_i^T \bar{u}_i \leq 0$. In both cases, we have $\dot{K} \leq 0$. Hence, $\dot{K}(v_i) \leq 0$ when $\|v_i\| = v_{max}$ and any trajectory $v_i(t)$ starting in the sublevel set $X_i = \{v_i: K(v_i) \leq (1/2)v_{max}^2\}$ cannot leave this set.

**Claim 2**: Given a collection of robots, with structural dynamics $\Sigma$, we have $\dot{H}_\lambda \leq -c_2 v^T v \leq 0$.

**Proof**:
$$\dot{H}_\lambda(p(t),v(t)) = \sum_i((\partial H_\lambda/\partial p_i)\dot{p}_i + (\partial H_\lambda/\partial v_i)\dot{v}_i) = \sum_i((\partial U_\lambda/\partial p_i)^T v_i + v_i^T u_i) \quad (25)$$

$$(\partial U_\lambda/\partial p_i)^T v_i + v_i^T u_i = \begin{cases} (\partial U_\lambda/\partial p_i)^T v_i & \text{if } v_i^T \bar{u}_i > 0 \text{ and } \|v_i\| \geq v_{max} \\ -c_2 v_i^T v_i & \text{otherwise} \end{cases} \quad (26)$$

In the first part of (26) we have

$$v_i^T \bar{u}_i > 0 \rightarrow -v_i^T \frac{\partial U_\lambda}{\partial p_i} - c_2 v_i^T v_i > 0 \rightarrow v_i^T \frac{\partial U_\lambda}{\partial p_i} < -c_2 v_i^T v_i \quad (27)$$

From (26) and (27) we have

$$\left(\frac{\partial U_\lambda}{\partial p_i}\right)^T v_i + v_i^T u_i \leq -c_2 v_i^T v_i \rightarrow \dot{H}_\lambda \leq -\sum_i c_2 v_i^T v_i \rightarrow \dot{H}_\lambda \leq -c_2 v^T v \leq 0 \quad (28)$$

So, the Hamiltonian energy of the system is non-increasing for all $(p,v)$.

**Theorem 1**: Consider a group of agents applying the FMP algorithm with $c_1, c_2 > 0$ and structural dynamics $\Sigma$. Assume that the initial kinetic energy of the system $K(v(0)) = 0$ and the inertia $J(p(0))$ is finite. Also assume $H_\lambda(p(0),v(0)) < \psi(d^*)$, then none of the agents can come closer than $d^*$ to each other.

**Proof**: Suppose $H_\lambda(p(0),v(0)) < \psi(d^*)$ and there is at least one pair of agents that their distance becomes less than $d^*$ at a given time $t_1 \geq 0$. This implies the collective potential of the particle system is at least $\psi(d^*)$: $H_\lambda(p(t_1),v(t_1)) \geq \psi(d^*)$. However, by Claim 2, $H_\lambda(p(0),v(0)) \geq H_\lambda(p(t_1),v(t_1))$ so we have:

$$H_\lambda(p(0),v(0)) \geq H_\lambda(p(t_1),v(t_1)), H_\lambda(p(t_1),v(t_1)) \geq \psi(d^*)$$
$$\rightarrow H_\lambda(p(0),v(0)) \geq \psi(d^*)$$

this contradicts with the assumption that $H_\lambda(p(0),v(0)) < \psi(d^*)$. Hence, none of the agents can come closer than $d^*$ at any time $t \geq 0$.

*D.2.2 Calculating the value of $d^*$*

Having Theorem 1 we are now ready to calculate the guaranteed minimum separation distance $d^*$ by the FMP algorithm. In this theorem we assume that $H_\lambda(p(0),v(0)) < \psi(d^*)$ and $K(v(0)) = 0$. We also assume the initial distance between the agents is equal or greater than $d$ (see section III.B). We know that in the 2D environment each agent can have at most 6 agents within its distance $d$ (otherwise the neighbors have to become closer to each other than d). This number grows to 12 agents in 3D case. We continue the proof for the 2D environment. The maximum initial Hamiltonian energy of the system is calculated as

$$H_\lambda(p(0),v(0)) = U_\lambda(p(0)) + K(v(0)) = V(p(0)) + c_1 J(p(0))$$
$$V^{max}(p(0)) = (1/2)\sum_i \sum_{j \neq i} \psi(d) = (1/2)\sum_i \sum_{j \neq i} \rho/3 \times (r-d)^3$$
$$< 6n \times \rho(r-d)^3/(2 \times 3) = n\rho(r-d)^3 \quad (29)$$
$$J(p_0) = (1/2)\sum_{i=1}^{n}\|p_i - \mathcal{T}_i\|^2 \leq \xi n/2 \quad (30)$$

where $\xi$ is the maximum Euclidean distance between the agents and their goal locations. From (29) and (30) we have

$$H_\lambda(p(0),v(0)) < n\rho(r-d)^3 + \xi n/2 \quad (31)$$

As we assumed $H_\lambda(p(0),v(0)) < \psi(d^*) = -\rho/3 \times (d^*-r)^3$ so we have

$$d^* = r - \sqrt[3]{3n(r-d)^3 + (3n\xi)/(2\rho)} \quad (32)$$

These calculations show if the initial distance between the agents is at least d and their initial velocity is zero then none of them become closer than $d^*$ using FMP algorithm. Using very similar calculations for the 3D environment we have

$$d^* = r - \sqrt[3]{6n(r-d)^3 + (3n\xi)/(2\rho)} \quad (33)$$

Having the value of $d^*$ the value of parameter $d$ can be calculated for 2D environment as

$$d = r + \sqrt[3]{(d^*-r)^3/(3n) + \xi/(2\rho)} \quad (34)$$

Applying (15) we have (35) for 2D and (36) for 3D.

$$d = d^* + \sqrt[3]{((9n-3) \times v_{max}^2 + 3n\xi)/(2\rho)} \quad (35)$$
$$d = d^* + \sqrt[3]{((18n-3) \times v_{max}^2 + 3n\xi)/(2\rho)} \quad (36)$$

Equation (36) and (15) are used in FMP algorithm for calculation of a communication distance $r$ that guarantees given separation distance $d^*$.

*D.3 Convergence analysis*

This section presents analysis that shows using the FMP algorithm all the agents converge to final positions and zero velocities. By Barbalat lemma we have $\lim_{t \to \infty} v(t) = 0$. Consequently, there exist a $T > 0$ such that $\|v_i(t)\| < v_{max} \quad \forall i, t \geq T$. Thus, from T onward, the system satisfies

$$\begin{cases} \dot{p}_i = v_i \\ \dot{v}_i = -\partial U_\lambda(p)/\partial p_i - c_2 v_i \end{cases} \quad (37)$$

which is a continuous time invariant system. Since from claim 2 we have $H(t) \leq H(0) \quad \forall t$, we know all signals are bounded. Hence LaSalle's theorem applies. We look for the largest invariant set on which $v(t) \equiv 0$. Setting $v(t) \equiv 0$ in the dynamics, we obtain $\partial U_\lambda(p)/\partial p_i = 0$. Hence, $p(t)$ approaches a stationary point of $U_\lambda$.

## IV. SIMULATION RESULTS

In this section, we present several simulation results of 2-D and 3-D FMP over three main benchmark problems. We compare the results with the most promising existing algorithm ORCA [7]. The Java implementation of the ORCA algorithm is used in these experiments [22]. We have implemented the FMP algorithm in Java, but our simulator is implemented in MATLAB. Both FMP and ORCA algorithms are called from Java in the MATLAB simulator to have a fair comparison. These experiments evaluate FMP algorithm over scalability issue, obstacle avoidance, deadlock avoidance, optimality of the produced motion and the ability to solve dense as well as 3D problems. The algorithms are compared over transition and execution time parameters. Transition time is the time that takes for agents to move from their initial positions to final positions applying the motions produced by the algorithm and the execution time is the offline execution time of the algorithm (time taken to solve the motion planning problem) which shows the computational complexity of the algorithm. We execute our implementation on a laptop running Windows 10, with an Intel Core i7-7700HQ 2.8GHz CPU and 16 GB RAM. The step-size $\Delta t$ is set to $0.02s$ and EndMaxD $= 0.05m$ for FMP in all the simulations. The videos accompanying this section are available in Extensions.

*A. Dense benchmark problems test*

In this Section we compare the performance of the FMP algorithm versus ORCA over three benchmark problems [23] with 100 agents represented in Fig. 5. The first case shows 100 agents whose goal is to move across a circle to antipodal position. The second and third cases show two different swap positions for 100 agents: mirror and diagonal. For both algorithms: in the circle example $d^* = 3\ m$ and $V_{max} = 15\ m/s$ and in both swap position examples $d^* = 5\ m$ and $V_{max} = 15\ m/s$. Note that as ORCA doesn't have any direct input parameter to set required separation distance, we set the input parameter $radius = d^*/2$ which yields the same result. We use the same method in other ORCA tests as well.

Table 1 compares the transition and execution times of FMP and ORCA algorithms over these three benchmark problems. As this table shows for all the test cases FMP greatly outperforms ORCA in terms of both transition and execution times.

To show the performance of the algorithms over dense problems we increase the problems' density on the second and third benchmarks in several steps. Therefore, the initial and final horizontal and vertical distance between neighboring points is decreased from $9.5m$ to $6m$ to shrink the problem. The results are compared in Table 1.

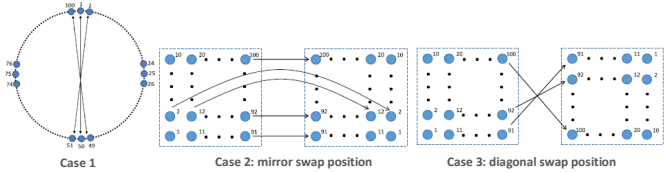

Fig. 5. Three benchmark motion planning problems with 100 agents

TABLE 1
TRANSITION AND EXECUTION TIME FOR THREE BENCHMARK PROBLEMS FOR FMP AND ORCA ALGORITHMS

| Example | initial distance (m) | FMP algorithm | | ORCA algorithm | |
|---|---|---|---|---|---|
| | | Transition time (s) | Execution time (ms) | Transition time (s) | Execution time (ms) |
| circle | - | 23.22 | 292 | 47.85 | 784 |
| mirror swap | 6 | 93.50 | 594 | no-result | no-result |
| | 6.5 | 49.52 | 364 | 597.25 | 11328 |
| | 7.5 | 36.06 | 273 | 460.75 | 7849 |
| | 8.5 | 32.27 | 273 | 166.92 | 2092 |
| | 9.5 | 29.49 | 272 | 76.44 | 1078 |
| diagonal swap | 6 | 38.28 | 287 | no-result | no-result |
| | 6.5 | 44.4 | 396 | 837.63 | 15956 |
| | 7.5 | 21.77 | 200 | 287.24 | 4883 |
| | 8.5 | 27.19 | 217 | 258.52 | 3808 |
| | 9.5 | 36.84 | 293 | 266.84 | 4317 |

As the results show the FMP algorithm has lower transition times especially in dense problems. ORCA, while claiming to be an optimal algorithm, in reality struggles with finding a time-optimal solution. This is most clearly shown in the original ORCA author's video post of 14 agents position exchange on a circle which is a highly packed scenario (minute 1:25 of the video [24]). The reason is that ORCA has to reduce the agents speed in order to keep its stability which causes long transition times especially in dense problems which is the problem of the focus of many applications. Also, as we expected and as is explained in Section III.D.1 although both algorithms have the same order of computational complexity, as calculation of the value of the repulsive function is the only computational effort that should be done for each neighbor at each step of the FMP algorithm, the execution time of the FMP is much lower than ORCA and this difference is greater in denser problems. It is also the result of a lower number of steps each agent takes to reach to their goals using the FMP algorithm. Note that transition time is a factor of the number of steps and the step size ($transition\ time = \#steps \times \Delta t$). As the step size for both algorithms is equal ($\Delta t = 0.02$ s), the algorithm that has lower transition time converges to its final solution in a fewer number of steps.

*B. Scalability tests*

We test the scalability of the FMP algorithm over the circle benchmark problem depicted in case 1 of Fig. 5. $d^*=5m$ and $V_{max} = 15\ m/s$ in this test. We increase the number of agents from 10 to 1000. Table 2 shows the execution and transition time for this test. As the results show the algorithm scales well for large number of agents. In addition, as each agent makes independent decisions, the algorithm is able to be parallelized efficiently by distributing the computations for agents across multiple processors. In these experiments we don't use the parallelization, but it can be added simply to later extensions. Fig. 6 shows three snapshots of the FMP algorithm for the circle benchmark problem with 1000 agents which shows that the agents move smoothly through the congestion that forms in the center.

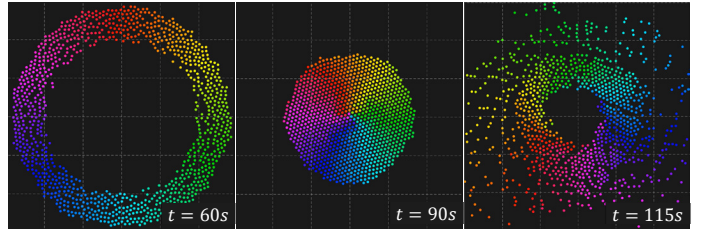

Fig. 6. Simulation of 1000 agents trying to move to antipodal positions at transition times: 60s, 90s, 115s. Agents smoothly move through the congestion that forms in the center. The full video is available as Extension.

TABLE. 2
TRANSITION AND EXECUTION TIME OF FMP OVER CIRCLE BENCHMARK PROBLEM WITH UP TO 1000 AGENTS AND WITHOUT PARALLELIZATION

| Number of agents | 10 | 100 | 250 | 500 | 1000 |
|---|---|---|---|---|---|
| Transition time (s) | 7.51 | 28.8 | 67.86 | 145.54 | 270.14 |
| Execution time (ms) | 28 | 245 | 1949 | 12402 | 72343 |

*C. Obstacle avoidance tests*

This section presents simulation results which show that the obstacle avoidance version of the FMP algorithm is able to avoid dynamic and static obstacles successfully. In these experiments $\acute{\rho} = \rho = 7.5 \times 10^6$ and $\acute{d}^* = 0.5\ \text{m} < d^*$. The first simulation shows four groups of 25 agents diagonally swapping their positions through a passage formed by the presence of four static obstacles. Fig. 7 shows four snapshots of the FMP algorithm solving this problem. It is numerically verified that no agent ever entered any of the four obstacles. The video that is available as Extension video shows how the algorithm successfully avoids dynamic obstacles as well as static ones.

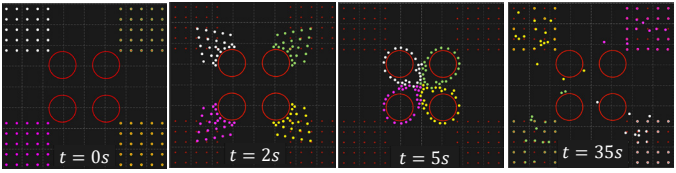

Fig. 7. Simulation of 4 group of 25 agents that moves with opposite directions through a narrow passage formed by presence of 4 static obstacles. Red dots represent initial and goal positions. Represented times indicating transition time. The full video is available as Extension.

### D. Random deadlock avoidance tests

In this simulation we generated 100 random test cases to compare the average performance of the ORCA and FMP algorithms over transition and execution times and the number of deadlocks occurring in each algorithm. Each test case includes 30 random points generated in a $[40m \times 40m]$ environment. $d^* = 5\ m$ and $V_{max} = 3\ m/s$. Poisson-disc sampling [25] is used for producing both initial and final random points that are packed but no closer to each other than the minimum distance $d$. Table 3 shows the result of this comparison.

TABLE. 3
COMPARISON OF FMP AND ORCA OVER 100 RANDOM TEST CASES

|  | FMP | ORCA |
|---|---|---|
| Number of deadlocks | 0 | 9 |
| Overall min separation distance (m) | 5.07 | 4.98 |
| Average Transition time (s) | 47.30 | 422.51 |
| Average Execution time (ms) | 99.24 | 1527.45 |

As is represented in Table 3, ORCA becomes stuck in deadlock in 9 percent of the problems. Deadlock is one of the drawbacks of most motion planning algorithms especially in packed problems. The deadlock problem in ORCA is discussed in [7, 26, 27, 28] and some deadlock avoidance versions have been proposed. Reference [7] explains the deadlock problem in a non-holonomic version of ORCA in details. Note that this deadlock issue is inherited from the original ORCA for holonomic agents [28]. ORCA experiences deadlock in packed scenarios in which the distance between goal locations is less than $4r$, with $r$ the radius of the agents [7]. In such scenarios, deadlock for ORCA happens when all the agents reach their goal locations except a few ones that are not able to reach their goal positions because their motion to the goal is blocked by other agents which are already in their final positions. An extension video shows the performance of FMP on a famous ORCA deadlock scenario (Fig. 8 in [7]). This paper presents mathematical analysis on the convergence of FMP (Section III.D.3) which proves that the algorithm doesn't have any livelock. Livelock is a condition that occurs when two or more agents continually change their positions in response to change in other agents and as a result none of the agents make any progress. As Table 3 shows, FMP provides lower transition and execution times and better guarantees minimum separation distance $d^*$. These results confirm the results reported in Section IV.A.

Another interesting result that the data presented in Tables 1,2,3 reveals is that in all the experiments, the execution time of FMP is always much less than its transition time. This means that FMP is able to do the calculations for each agent while it is transiting toward its goal location.

### E. Time-optimality test

This section presents simulation results for evaluation of time-optimality of the motion produced by the FMP algorithm. Equation (38) presents a theoretical lower bound for the optimal transition time.

$$Lbt_{opt} = lower\ bound\ of\ t_{opt} = Max_{i=1}^{n}(\|\mathcal{F}_i - \mathcal{I}_i\|/v_{max}) \quad (38)$$

$Lbt_{opt}$ shows the time that takes for all the agents to reach their goal locations if they all go straight toward their goal at their maximum speed. This value is a lower bound of the optimal transition time because it doesn't consider the time agents need to avoid collision with each other. Although $Lbt_{opt}$ is not the optimal transition time, it still is able to give us a good estimation of how far the FMP transition time is from the optimal value.

In this simulation agents change their formation from a circle with 28 nodes to a double-circle. Fig. 8 shows snapshots of initial and final formations. The initial and final distance between nodes decreased from $12m$ to $1m$ to make a range of easy to hard problems. The parameters are set as follows: $d^* = 0.4m$, $v_{max} = 10\ m/s$. Fig. 9 compares the quality of the motions produced by FMP and ORCA with the lower bound of the optimal motion ($Lbt_{opt}$).

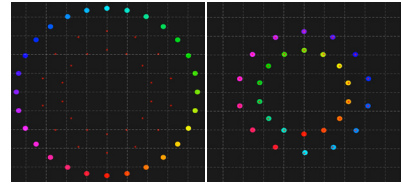

Fig. 8. Snapshots of initial and final agents' formations in time-optimality test. Agents color shows how initial and final nodes are assigned to each other.

The optimal transition time lie somewhere between the $Lbt_{opt}$ and FMP lines in this figure. As represented in this figure, in all the easy to hard testcases, the transition time of the motion produced by FMP is very close to $Lbt_{opt}$ and is even closer to the optimal transition time consequently. ORCA produces near to optimal transition time for low density problems but as the problem becomes harder its transition time increases strongly.

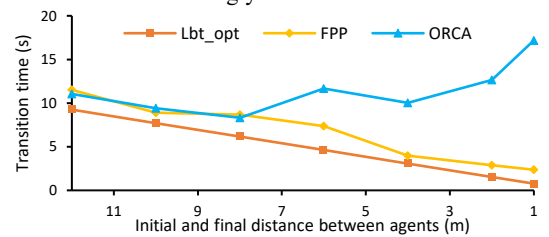

Fig. 9. Comparison of FMP and ORCA produced motions transition times with theoretical lower bound of optimal transition time ($Lbt_{opt}$).

### F. 3D FMP

The 3D simulation of the FMP algorithm is presented in this section. Fig. 10 shows the consecutive snapshots of 3D simulation of FMP for 100 agents. Each agent represents a UAV moving in 3D. The initial and final state of the agents is chosen at random using poisson-disc sampling. The parameters are set as follows: $d^* = 3m$, $v_{max-horizental} = 9\ m/s$, $v_{max-up} = 3\ m/s$, $v_{max-down} = -6\ m/s$. As the figure shows the agents reach their goal locations in few seconds and the minimum separation distance guarantee is numerically verified as is depicted in Fig. 11. The full video of this experiment and a few more 3D examples is available as Extension.

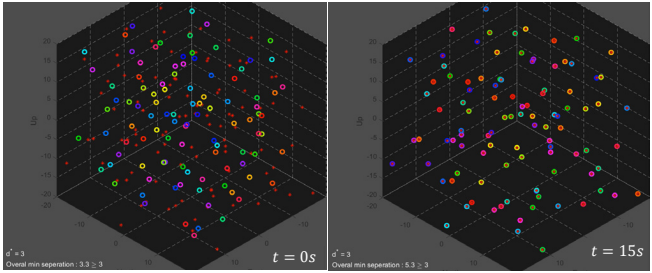

Fig. 10. Simulation of 100 3D agents with random initial and goal locations using 3D version of FMP as the motion planner. Red dots represent the goal locations. $t$ indicates transition time. The full video is available as Extension.

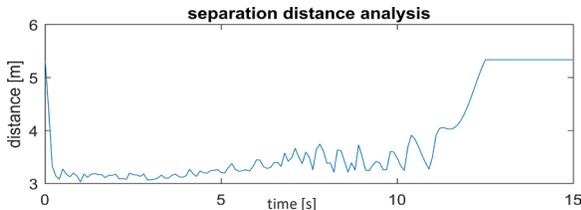

Fig. 11. Minimum distance between all agents in [0, 15s] time interval in 3D version of FMP algorithm

## IV. CONCLUSION

This paper presents FMP, a real-time force-based motion planning algorithm. We show that the algorithm addresses several practical issues in motion planning of a multi-agent system. First, it is a completely distributed algorithm in which each agent only requires knowledge of the relative positions of neighboring agents. Second, it has very low computational and communication overhead over the agents. Third, the algorithm scales well to very large agent teams involving thousands of agents. Fourth, using this algorithm agents produce near to time-optimal motions that have much lower transition time in comparison to other famous collision avoidance algorithms. Fifth, the algorithm enables agents to effectively avoid both dynamic and static obstacles in the environment. Sixth, it is able to produce deadlock free motions even in dense and packed scenarios. Seventh, the algorithm covers the movement of agents in both 2D and 3D environments. Eight, it has solid mathematical background that not only guarantees production of collision free motions but also guarantees keeping the minimum required separation distance between the agents. In addition, it guarantees the convergence of the algorithm. Finally, it supports user preferences such as maximum allowed velocity of agents and minimum required separation distance between them. For future work, we are investigating solutions to produce more time-optimal motions leading to lower transition time. Also, expanding this algorithm to take into account agents with more complicated kinematic constraints can be considered as another future work for this research.

## APPENDIX

This paper has supplementary downloadable material available at https://drive.google.com/open?id=1Od3DK8_4YCIWoQgMfgpfglm_1sMBB6NF. This includes an MP4 video showing simulations from Section IV including scalability, obstacle avoidance and deadlock avoidance tests and the ability to solve dense problems.

10[24] J. Alonso-Mora, A. Breitenmoser, M. Rufli, R. Siegwart and P. Beardsley, (2018, March 28), Optimal Reciprocal Collision Avoidance for Multiple Non-Holonomic Robots, Retrieved from https://www.youtube.com/watch?v=s9lvMvFcuCE.

[25] R. Bridson, Fast Poisson disk sampling in arbitrary dimensions. In SIGGRAPH sketches, p. 22, 2007.

[26] A. Giese, D. Latypov, and N.M. Amato, 2014, May. Reciprocally-rotating velocity obstacles. In Robotics and Automation (ICRA), IEEE International Conference on (pp. 3234-3241). 2014.

[27] S. A. Khan, M.J. Yasar Ayaz, S. O. Gillani, M. Naveed, A.H. Qureshi, and K.F. Iqbal, Collaborative Optimal Reciprocal Collision Avoidance for Mobile Robots. International Journal of Control and Automation, 8(8), pp.203-212, 2015.

[28] J. Alonso-Mora, A. Breitenmoser, M. Rufli, P. Beardsley, and R. Siegwart, Optimal reciprocal collision avoidance for multiple non-holonomic robots. In Distributed Autonomous Robotic Systems, pp. 203-216, Springer, Berlin, Heidelberg, 2013.